\documentclass[letterpaper, 10 pt, conference]{ieeeconf}  
\usepackage{cite}
\usepackage{graphicx}
\usepackage{tcolorbox}
\usepackage{xcolor}
\usepackage{hyperref}
\usepackage[]{footmisc}
\usepackage{adjustbox}
\usepackage{booktabs,graphicx,makecell,siunitx,eqparbox}
\usepackage{tabularx}
\usepackage{footnote}
\usepackage{balance}
\usepackage{svg}
\usepackage{caption}
\usepackage{subcaption}
\usepackage{amssymb}
\usepackage{amsmath}
\usepackage{algorithm}
\usepackage{algpseudocode}
\makesavenoteenv{tabular}

\IEEEoverridecommandlockouts                              

\title{\LARGE \bf
Language-Guided Object-Centric Diffusion Policy for \\
Generalizable and Collision-Aware Manipulation
}

\author{Hang Li$^{*,1}$, Qian Feng$^{*,1,2}$,  Zhi Zheng$^{1,2}$, Jianxiang Feng$^{1,2}$, Zhaopeng Chen$^{2}$, Alois Knoll$^{1}$ 
\thanks{*: Equal Contributions, \{hang1.li, qian.feng\}@tum.de.}
\thanks{$^{1}$Technical University of Munich}
\thanks{$^{2}$Agile Robots SE}
}

\begin{document}

\maketitle
\thispagestyle{empty}
\pagestyle{empty}

\begin{abstract}
Learning from demonstrations faces challenges in generalizing beyond the training data and often lacks collision awareness. This paper introduces Lan-o3dp, a language-guided object-centric diffusion policy framework that can adapt to unseen situations such as cluttered scenes, shifting camera views, and ambiguous similar objects while offering training-free collision avoidance and achieving a high success rate with few demonstrations. We train a diffusion model conditioned on 3D point clouds of task-relevant objects to predict the robot's end-effector trajectories, enabling it to complete the tasks. During inference, we incorporate cost optimization into denoising steps to guide the generated trajectory to be collision-free. We leverage open-set segmentation to obtain the 3D point clouds of related objects. We use a large language model to identify the target objects and possible obstacles by interpreting the user's natural language instructions. To effectively guide the conditional diffusion model using a time-independent cost function, we proposed a novel guided generation mechanism based on the estimated clean trajectories. In the simulation, we showed that diffusion policy based on the object-centric 3D representation achieves a much higher success rate (68.7\%) compared to baselines with simple 2D (39.3\%) and 3D scene (43.6\%) representations across 21 challenging RLBench tasks with only 40 demonstrations. In real-world experiments, we extensively evaluated the generalization in various unseen situations and validated the effectiveness of the proposed zero-shot cost-guided collision avoidance.
\end{abstract}

\section{INTRODUCTION}
For robots to perform manipulation tasks in daily life, they should be able to adapt to complex and dynamic environments, ensuring safe and efficient task completion. Imitation learning offers a method for robots to acquire skills quickly. However, a key limitation is that these skills often struggle to generalize beyond the training data to new scenarios. Some research works \cite{brohan2023rt2, openvla} aim to achieve skill generalization by training large models for robots. Yet, this approach demands an extensive amount of training data. A highly desired goal is to enable skill generalization to unseen situations using only a small amount of data while also addressing obstacle avoidance and safety concerns.

Recently, diffusion models have shown great potential in the field of robotic manipulation~\cite{janner2022planning, ma2024hierarchical, gdp}. 
In the realm of imitation learning, diffusion-based methods~\cite{IHB, chi2023diffusionpolicy, Ze2024DP3} have demonstrated strong capabilities towards learning complex manipulation tasks. Compared to traditional imitation learning algorithms, diffusion models offer the advantages of stable training, high-dimensional output spaces, and the ability to capture the multi-modal distribution of actions~\cite{chi2023diffusionpolicy}. Nevertheless, their performance is still limited when the testing scenes differ from the training scenes in terms of background changes, camera view shifts, multiple similar target objects, the presence of obstacles, and so on.

To address the challenges of generalization and safe obstacle avoidance, we propose Lan-o3dp, a language-guided, object-centric 3D diffusion policy framework that is generalizable and collision-aware. The diffusion model is trained to predict robot end effector trajectories conditioned on the segmented 3D point clouds of task-relevant objects by denoising random noise into a coherent action sequence. By filtering out task-irrelevant visual information and preserving only the 3D data of task-relevant objects from a calibrated camera, policy learning becomes more data-efficient, while significantly enhancing generalization performance. 

During inference, while the point clouds of target objects serve as visual observations for the diffusion policy, the segmented point clouds of obstacles are transformed into a cost function, which is incorporated into the denoising step for iterative optimization. This ensures that the generated trajectory completes the task while remaining collision-free. The cost function is constructed using the location and geometric information of obstacles obtained from the segmented point clouds. Since the cost function is independent of the diffusion time steps, we propose a novel guided sampling method—calculating the cost of the estimated clean trajectory rather than the noisy trajectories, to achieve more effective guidance. We design the cost function that enables the robotic arm's end-effector to avoid obstacles of different shapes and quantities by continuously updating the closest point between the end-effector and the obstacles in real time.

We have observed that object-centric 3D representations offer benefits in environments with obstacles. They not only prevent policy failures that might result from changes in visual observations due to the presence of obstacles but also allow us to obtain precise positional information on obstacles from calibrated point clouds. Moreover, by adjusting the camera viewpoint, we can observe objects that are otherwise occluded by obstacles.

We leverage open-set segmentation \cite{GroundingDino, SAM, Cutie} to obtain segmented point clouds of target and obstacle objects based on their names. This further enhances the framework's generalization: We can specify the target object through language in visually ambiguous scenarios, such as those with multiple similar objects. This also enables open-vocabulary manipulation for objects with similar geometries and open-vocabulary collision avoidance for novel obstacles. We use large language models to parse the required skills, target objects, and obstacles from human natural language instructions. The language model generates code to invoke detection, segmentation, and tracking modules to obtain the necessary point clouds and executes the policy to complete the task, which enhances the system's flexibility.



We demonstrate the effectiveness of our proposed methods against state-of-the-art diffusion-based methods in 21 RLBench \cite{james2019rlbench} simulation tasks and further extensively evaluate the generalization in various unseen situations and validate the effectiveness of
the proposed zero-shot cost-guided collision avoidance in real-world experiments.


In summary, our main contributions are:
\begin{enumerate}
\item We propose Lan-o3dp, a language-guided object-centric diffusion policy framework that generalizes across diverse aspects such as background changes, camera view shift, and even scenes of multiple similar objects while offering training-free collision avoidance and data efficiency.
\item We introduce a novel guidance mechanism in the diffusion sampling stage, which can more effectively guide the generation with time-independent cost.
\item We show Lan-o3dp can interpret high-level human natural language instruction and achieve the open-set manipulation for objects with similar geometries and open-set collision avoidance for unknown obstacles with various shapes and quantities.
\item The proposed method is evaluated in both simulation and real-world experiments, demonstrating the effectiveness and universality compared to the baselines.
\end{enumerate}
\section{RELATED WORK}
\subsection{Diffusion Models in Robotics}
The diffusion model is a type of probabilistic generative model that learns to generate new data by progressively applying a denoising process on a randomly sampled noise. 
Learning based robotic grasping~\cite{Liang_2019,Qian2022FFHNet,burkhardt2024multifingered, feng2024dexgangrasp} and manipulation skills~\cite{zhu2023groot,liang2024skilldiffuser, Bimanual, Surgical} are longstanding problems. 
Due to its advantage of stable training and impressive expressiveness, diffusion model has been applied in several robotic fields such as reinforcement learning~\cite{ajay2023conditional, janner2022planning, adaptdiffuser}, imitation learning~\cite{chi2023diffusionpolicy,Ze2024DP3}, grasp synthesis~\cite{barad2023graspldm,weng2024dexdiffuser} and motion planning~\cite{urain2023se3diffusionfields,MPD,EDMP}.
In this work, we utilize object-centric 3D representation to maximize the generalization ability of the trained policy based on imitation learning via diffusion model and introduce a novel guided diffusion mechanism for obstacle avoidance.


\subsection{Object-Centric Representation Learning} 
Object-centric representations have been widely studied to reason about modular visual observations in the robotic field.
In robotics, researchers commonly use 6D poses~\cite{pose1,pose2,pose3}, bounding boxes~\cite{wang2019deep,devin2017deep} or segmented masks~\cite{sdmaskrcnn} to represent objects in a scene. These representations are limited to known object categories or instances. 
Recent progress in open-world visual recognition has led to the development of substantial models across various domains, including object detection~\cite{GroundingDino}, object segmentation~\cite{SAM}, and video object segmentation~\cite{cheng2022xmem}. Groot\cite{zhu2023groot} trains a transformer policy using segmented 3D objects. However, Groot uses a Segmentation Correspondence Model to identify the target object and cannot handle scenes with multiple similar objects. We use open vocabulary segmentation, which allows for a more convenient specification of target objects through language. 

\subsection{Language models for robotics}
Large language models (LLMs) possess powerful language comprehension abilities and a wealth of common knowledge. As a result, they can be effectively utilized to understand human instructions and to plan robotic tasks at a high level. Code as Policies\cite{CodeasPolicies} explores using a code writing LLM to generate robot policy code based on language commands. Voxposer\cite{huang2023voxposer} plans the task and generates code by an LLM to compose value maps for zero-shot manipulation. SayCan\cite{saycan} utilizes an LLM to select skills from a library of pre-trained skills. Many works also explore using LLMs to write reward functions for training robotic skills~\cite{reward1, reward2, Eureka}.
In addition, With the help of pre-trained open vocabulary vision language model~\cite{GroundingDino, OWL-ViT, yolo-world}, the robot can ground the user's instruction to the real world and accomplish various and complex tasks~\cite{huang2023voxposer,stone2023openworld,brohan2023rt2}.
In this work, we apply language models to select the desired policy and extract objects and obstacles from the user's instructions to obtain an object point cloud with the vision language model.
\section{Method}
\vspace{3pt}
\begin{figure*}[t]
\vspace{3pt}
\centering
\vspace{3pt}
\includegraphics[width=0.99\textwidth]{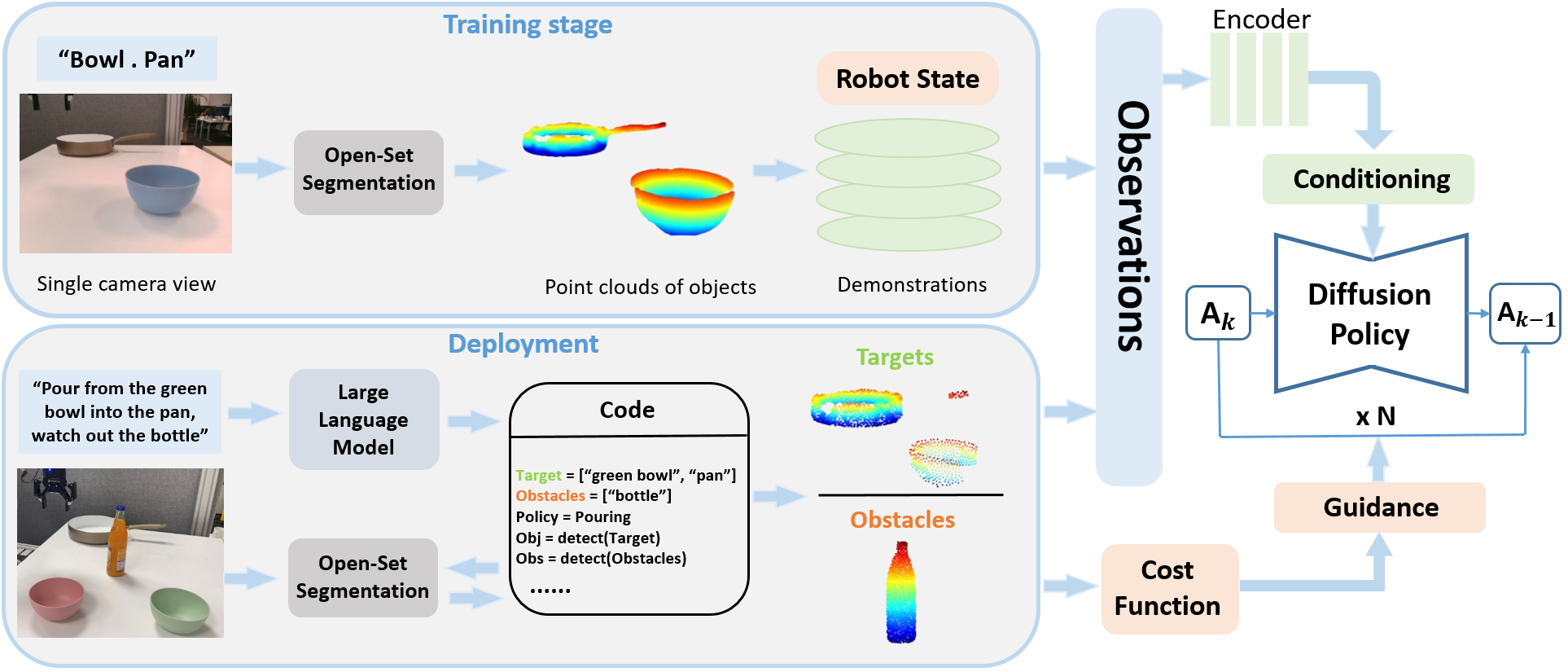}
\caption{An illustration of the proposed pipeline of Lan-o3dp. We use open-set segmentation to obtain the point clouds of objects. At the training stage, the visual observations in the demonstrations we collected only contained point clouds of objects relevant to the task. During deployment, a large language model is employed to decompose users' instructions into target objects and obstacles and select the corresponding policy given a set of trained policies. Target objects are used as visual observation for the model, while obstacles are transformed into a cost function to guide the model in generating collision-free trajectories.}
\label{fig:image1}
\vspace{-5pt}
\end{figure*}

\subsection{Problem Formulation and Preliminaries}
In this work, we want to address the generalization problem of learning visuomotor policy and introduce training-free collision awareness at test time. We explore the visual observation conditioning and the cost-guided generation of diffusion policy to solve these problems. 

\textbf{Diffusion policy visual conditioning:} 
Diffusion policy\cite{chi2023diffusionpolicy} uses DDPM to model the action sequence \( P(A_t \mid O_t) \). Wherein, \( A_t = \{ a_t, \ldots, a_{t+n} \} \) is the predicted next \( n \) action steps, which is a sequence of end-effector poses. The prediction horizon $n$ indicates that the diffusion policy predicts a trajectory over a shorter horizon in the closed loop instead of the entire trajectory. \( O_t = \{ V_t, S_t \} \) represents the visual observation \( V_t \) and robot states observation \( S_t \). The observation features are fused to the policy network through FiLM \cite{FiLM}. While diffusion policy and its variant 3D diffusion policy \cite{Ze2024DP3} takes simple 2D images or 3D point clouds of the whole scene as visual observation, we employ the segmented point clouds of task-relevant objects from a calibrated camera for policy learning. By eliminating redundant visual information and retaining only task-relevant data, policy learning becomes more data-efficient, and the model can minimize the negative effects of scene changes, thereby improving generalization performance.

\textbf{Guided sampling formulation:}
The diffusion model is trained to predict the added noise $\epsilon (O, A_k, k)$ at each diffusion timestep \( k \). During the reverse diffusion process, it gradually denoises a Gaussian noise to a smooth noise-free trajectory. The reverse process step is $A_{k-1} = \mu_k + \sigma_k z$, where $\mu_k = \frac{1}{\sqrt{\alpha_k}} \left( A_k - \frac{1-\alpha_k}{\sqrt{1-\overline{\alpha}_k}} \varepsilon \right)$, $z \sim \mathcal{N}(0, I)$, $\alpha_k \in \mathbb{R}$ and $\overline{\alpha}_k := \prod_{s=1}^k \alpha_s$ predefined scheduling parameters. Much prior work has explored guided sampling of the diffusion model. At the inference stage, guidance $g_k = \nabla_{A_k} D$ as a gradient term with respect to $A_{k}$ is added to the model's predicted mean such that each denoising step becomes:\begin{equation} A_{k-1} = \mu_k - \rho g_{k} + \sigma_k z \label{eq:1} \end{equation}, where $\rho$ is a scaling factor to control the effect of guidance. In this work, we model newly emerged obstacles in the scene as a cost function to guide the model in generating collision-free trajectories. 


\subsection{Approach}
\textbf{Training stage:}
Figure \ref{fig:image1} shows our pipeline, as shown in the training stage, we leverage open vocabulary segmentation, which is a combination of a vision language model (VLM), Segment Anything Model (SAM)~\cite{SAM} and a video object segmentation model~\cite{cheng2022xmem} to acquire real-time masks of the target objects and map these masks onto the point clouds given the object names. The demonstrations contain the point clouds of objects, robot states, and corresponding end-effector trajectories.



\textbf{Language guided deployment:}
During the deployment phase, the trained policies are applicable to different scenarios. Given a set of trained policies, a large language model is used to decompose the user's commands into policy, target objects, and obstacles. Similarly, open vocabulary segmentation is used to obtain the point cloud of the target objects and obstacles in each frame. The point cloud of target objects is subsequently inputted as an observation into the trained policy, while the point cloud of obstacles is processed and transformed into a cost function to guide the trajectory generation toward collision-free areas.

\textbf{Cost guided generation:}
In the field of robotics, many guided sampling techniques rely on reward models~\cite{janner2022planning, adaptdiffuser}, which are, however, often difficult to obtain. We choose to use a flexibly constructed cost function instead. To construct an effective cost function utilizing the 3D point cloud information of obstacles, we compute the Euclidean distance between each waypoint $\{a_0, \ldots, a_T\}$ in the generated action sequence $A_k$ and the obstacle point $C_{ob}$ that is closest to the end-effector. By setting a short action horizon and updating the closest point on the obstacles before each generation in the closed loop, we can avoid obstacles of different shapes and quantities. Calculating the closest point only once before the diffusion generation process can significantly reduce the computational load, ensuring that the guidance does not slow down the diffusion generation process.

As previously mentioned, most guided sampling methods~\cite{EDMP, MPD} calculate the cost/distance $D(A_k, C_{ob})$ of each intermediate action $A_k$ generated during the reverse diffusion process and compute the gradient $g_k = \nabla_{A_k} D(A_k, C_{ob})$. However, we found this kind of guidance can not provide sufficient influence for diffusion policy conditioned on observations. A cost function independent of the timestep $k$ of the diffusion process becomes less meaningful for noisy trajectories, especially in the early stages of the denoising process. Consequently, the cost of noisy trajectories struggles to provide effective guidance. Unlike previous methods~\cite{EDMP, MPD}, refer to FreeDoM \cite{FreeDoM}, we calculate the cost at each step based on the estimated $A_{0|k}$, an estimated clean trajectory. 
\begin{equation}
A_{0|k} := \mathbb{E}[A_0 | A_k] = \frac{A_k - \sqrt{1 - \overline{\alpha}_k} \epsilon_\theta (A_k)}{\sqrt{\overline{\alpha}_k}} \quad \text{\cite{DDPM}} \label{eq:2}
\end{equation}
We calculate the distance of $A_{0|k}$ estimated from $A_{k}$ at each timestep and use this cost to compute the gradient with respect to $A_{k}$, that is $\nabla_{A_k} D(A_{0|k}, C_{ob})$. Therefore, the equation \ref{eq:1} becomes:
\begin{equation} A_{k-1} = \mu_k - \rho\nabla_{A_k} D(A_{0|k}, C_{ob}) + \sigma_k z \label{eq:3} \end{equation} 

If any distance $D(a_i, C_{ob})$ in $D(A_{0|k}, C_{ob})$ is shorter than a safety critical distance $Q^*$ which is determined by the size of grasped object, a non-zero gradient will be assigned to the corresponding waypoint. In real-world experiments, we only consider the distance of x and y coordinates.
\begin{equation}
Gradient = 
\begin{cases}
\nabla D(a_i, C_{ob}), & \text{if } D(a_i, C_{ob}) \leq Q^* \\
0, & \text{if } D(a_i, C_{ob}) > Q^*
\end{cases}
\end{equation}

The proposed algorithms are shown in \textbf{Algo} \ref{alg:1} and \textbf{Algo} \ref{alg:2} within DDPM \cite{DDPM} and DDIM \cite{DDIM} sampling, respectively. 

\begin{algorithm}
\caption{Cost guided diffusion sampling (DDPM), given a diffusion model $\epsilon_\theta$, cost/distance measurement $D(x, y)$, current end-effector pose $P_{ee}$, point clouds of obstacles $P_{ob}$, gradient scale $\rho$ and $z \sim \mathcal{N}(0, I)$}
\label{alg:1}
\begin{algorithmic}[1] 
\State $C_{ob} \gets \arg\min_{p \in P_{ob}} \| p - P_{ee} \|_2$ \Comment{find the closest point $C_{ob}$ on obstacle}
\State $A_T \gets \text{sample from } \mathcal{N}(0, I)$
\For{$k = T$ to $1$}
    \State $\mu_k \gets \frac{1}{\sqrt{\alpha_k}} \left( A_k - \frac{1-\alpha_k}{\sqrt{1-\overline{\alpha}_k}} \epsilon_\theta \right)$
    \State $A_{k-1} \gets \mu_k + \sigma_k z$ 
    \vspace{5pt}
    \State $A_{0|k} \gets \frac{A_k - \sqrt{1 - \overline{\alpha}_k} \epsilon_\theta (A_k)}{\sqrt{\overline{\alpha}_k}} \quad$
    \State $A_{k-1} \gets A_{k-1} - \rho\nabla_{A_k} D(A_{0|k}, C_{ob})$
\EndFor
\State \text{\textbf{Return} $A_0$} 
\end{algorithmic}
\end{algorithm}

\vspace{-10pt}

\begin{algorithm}[h]
\caption{Cost guided diffusion sampling(DDIM)}
\label{alg:2}
\begin{algorithmic}[1] 
\State $C_{ob} \gets \arg\min_{p \in P_{ob}} \| p - P_{ee} \|_2$ 
\State $A_T \gets \text{sample from } \mathcal{N}(0, I)$
\For{$k = T$ to $1$}
    \State ${A}_{0|k} \gets \frac{A_k - \sqrt{1 - \overline{\alpha}_k} \epsilon_\theta (A_k)}{\sqrt{\overline{\alpha}_k}} \quad$
    \State $A_{k-1} \gets \sqrt{\alpha_{k-1}}{A}_{0|k} + \sqrt{1-\alpha_{k-1}-\sigma_k^2}\epsilon_\theta + \sigma_kz$
    \State $A_{k-1} \gets A_{k-1} - \rho\nabla_{A_k} D({A}_{0|k}, C_{ob})$
\EndFor
\State \text{\textbf{Return} $A_0$} 
\end{algorithmic}
\end{algorithm}
\vspace{-5pt}

\section{Experiment}
\label{sec:Experiments}
In simulation experiments, we demonstrate that the object-centric 3D representation is more effective for diffusion policy than simple 2D and 3D scene representation. In real-world experiments, we show that our method has strong generalization capabilities in unseen environments, and the proposed cost-guided generation can effectively avoid language-specified obstacles.

\subsection{Simulation Experiments}
\begin{figure}[t]
\centering
\vspace{5pt}
\includegraphics[width=0.48\textwidth]{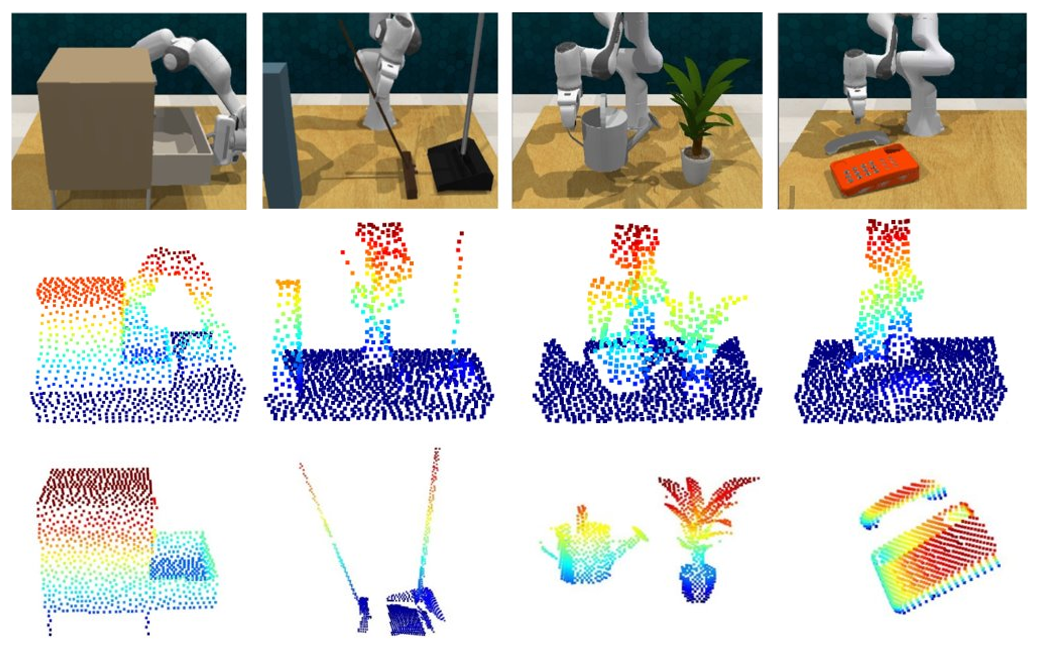}
\caption{Visualization of some simulation tasks. We use a single front camera to keep consistent with the real world. The top and middle row visualize the observations of two baselines: 2D RGB images and 3D point clouds of the scene from the front camera. The bottom row shows visualizations of our method of object-centric 3D point clouds.}
\label{fig:image2}
\vspace{-10pt}
\end{figure}

We conduct simulation experiments in RLBench to evaluate the success rate of proposed Lan-o3dp compared with two baselines, namely diffusion policy \cite{chi2023diffusionpolicy} and 3D diffusion policy \cite{Ze2024DP3}, aiming to demonstrate that 3D object-centric representation provides superior performance over simple 2D and 3D scene representations for diffusion policy. 


\paragraph{\textbf{Tasks}} We use only the front camera to evaluate the data efficiency to maintain consistency with real-world experiments. The key criterion for selecting tasks is that sufficient information to complete the task must be observable from the front camera. We select 21 challenging RLBench tasks covering manipulation, pick-and-place, single object, and multiple objects. 


\paragraph{\textbf{Demonstration and data processing}} We collect 40 demonstrations per task to keep consistency with the real world. Examples are shown in figure \ref{fig:image2}. We set the camera resolution to 120x120, and we extract the task-related object point cloud for our method. We downsample the object-centric point clouds to 256 points; For the baselines, we maintain their original processing methods \cite{Ze2024DP3} \cite{chi2023diffusionpolicy}. Each demonstration of every task includes variations, such as position changes of the objects. 

\paragraph{\textbf{Metric and results}} We design our model on top of CNN-based diffusion policy. We train 500 epochs for each task, evaluate 20 episodes every 50 epochs, and then compute the average of the highest 5 success rates. The episodes for evaluation also have variations. As shown in figure \ref{fig:right}, our method with object-centric representation achieves a significantly higher success rate, reaching an overall 68.7\% across 21 RLBench tasks. In contrast, the baseline methods—Diffusion Policy and 3D Diffusion Policy—that use simple 2D and 3D scene representations achieved only 39.3\% and 43.6\%, respectively. Through figure \ref{fig:left} we can further observe that our method achieves a higher number of tasks in the high success rate range, and only a few tasks have a low success rate.

\paragraph{\textbf{Ablation study}} We conducted additional ablation studies to examine our design choices across 7 RLBench tasks. Specifically, we evaluated three point cloud encoders: the PointNet encoder \cite{pointnet}, the MLP encoder \cite{Ze2024DP3}, and the MLP encoder with a residual connection (ours). Additionally, we assessed two learning objectives—"epsilon" prediction and "sample" prediction. Table \ref{tab:ablation} shows that the MLP Encoder-a simple multilayer perceptron-with a residual connection has the highest success rate, and the ``sample'' prediction slightly outperforms the ``epsilon'' prediction.

\begin{table}[ht!]
\vspace{-10pt}
\centering
\captionsetup[table]{skip=5pt}
\caption{Ablation studies on 7 RLBench tasks}
\vspace{-5pt}
\begin{center}
\label{tab:ablation} 
\begin{adjustbox}{width=0.8\linewidth}
\begin{tabular}{c|c}
\toprule [1.35pt]
Encoder / Prediction type &  \makecell{Average Success Rate}\\
\midrule
MLP \cite{Ze2024DP3} / Sample & ${64.1\%}$ \\ 
PointNet \cite{pointnet} / Sample  & $14.9\%$ \\ 
MLP+Residual / Sample  & $\textbf{68.8\%}$ \\ 
\cmidrule(lr){1-2} 
MLP+Residual / Epsilon  & $65.7\%$ \\ 
\bottomrule
\end{tabular}
\end{adjustbox}
\end{center}
\vspace{-15pt}
\end{table}

\begin{figure}[t]
\vspace{3pt}
\centering
\vspace{3pt}
\includegraphics[width=0.48\textwidth]{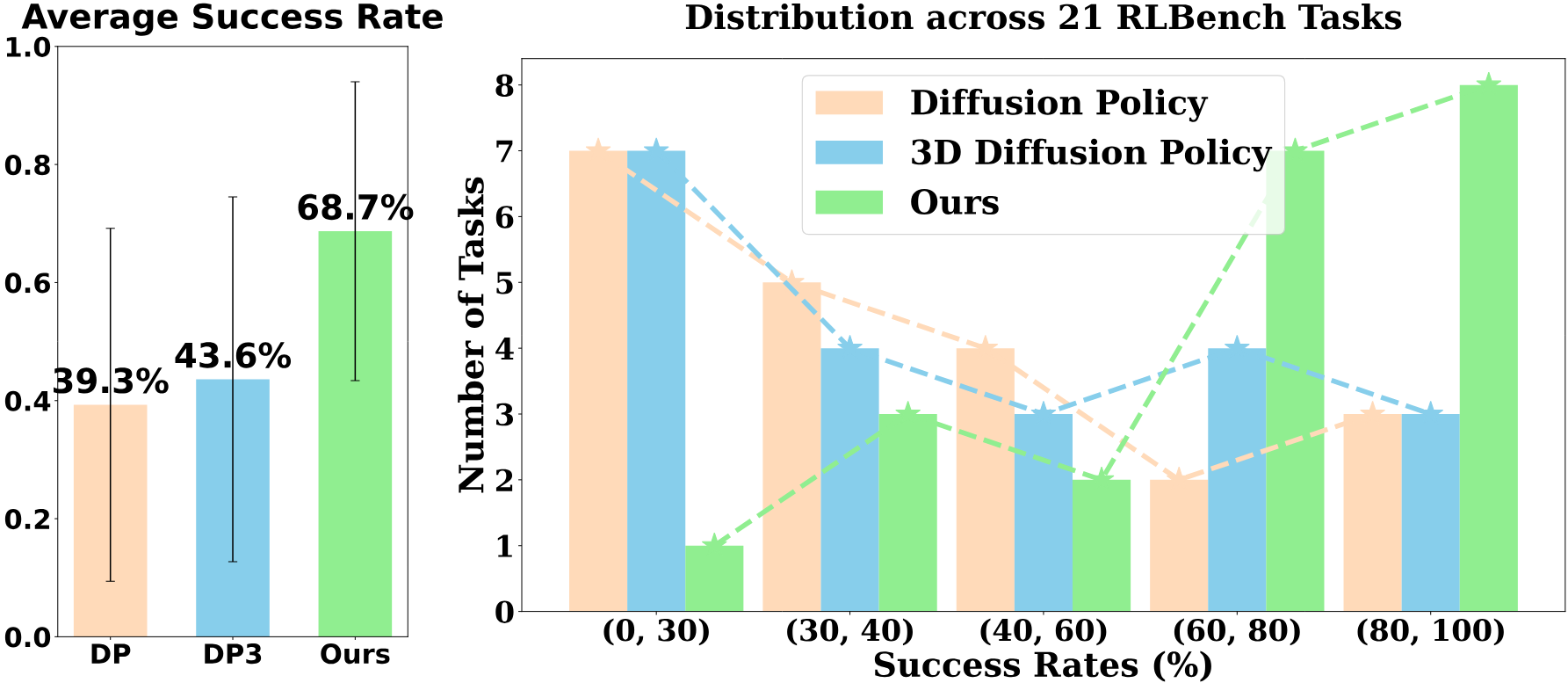}
\caption{Simulation results.
    (a) The average success rates over all 21 RLBench tasks.
    (b) The distribution of success rates. Our method with object-centric 3D representation achieves a higher average success rate and has a larger number of tasks in the 60-100\% success rate range.}
\label{fig:image3}
\vspace{-15pt}
\end{figure}


\subsection{Real World Experiments}
\subsubsection{Experiment Setup}
\begin{figure}[t]
\vspace{3pt}
\centering
\vspace{3pt}
\includegraphics[width=0.4\textwidth]{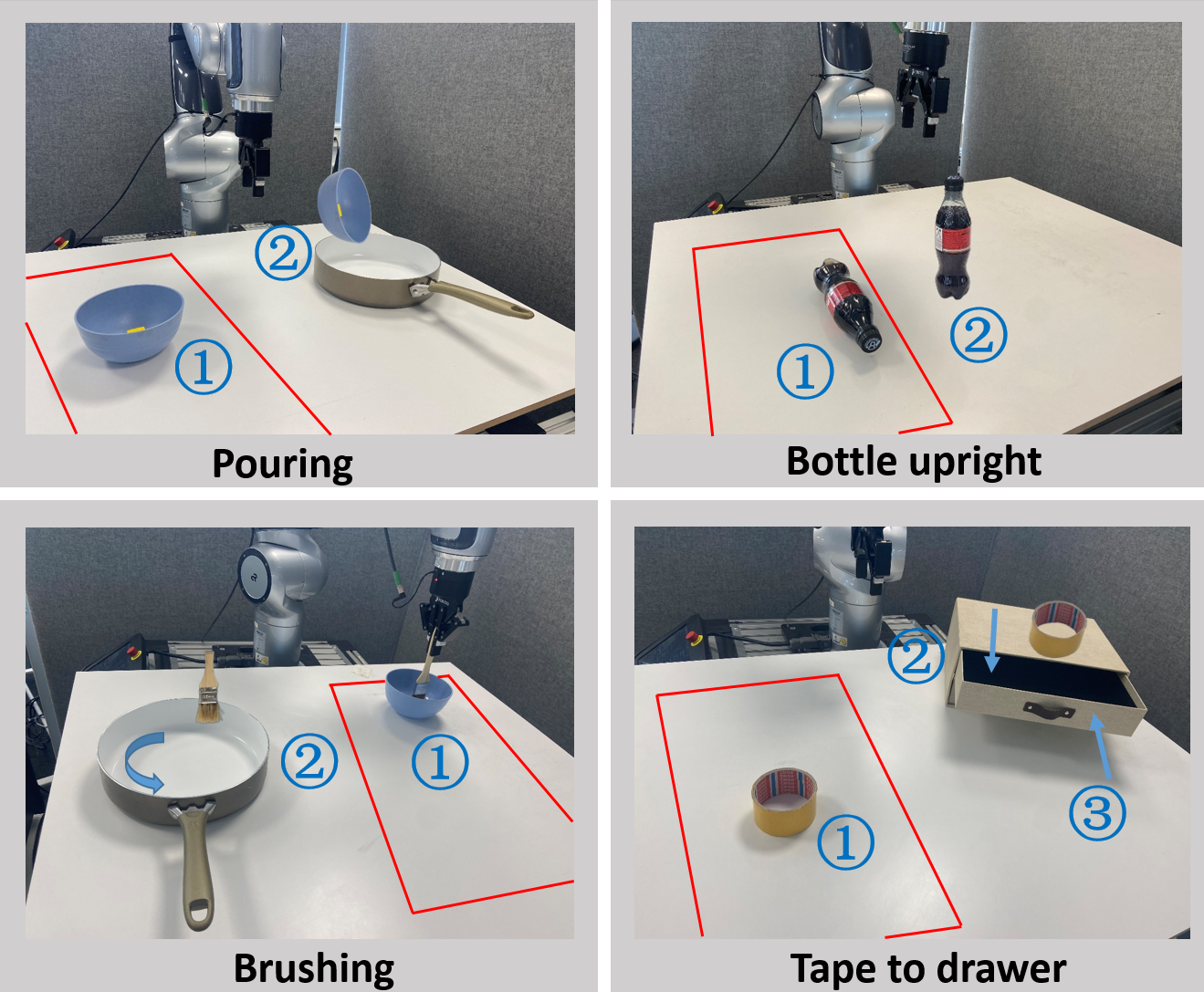}
\caption{Four tasks in real-world experiments: pouring, bottle upright, brushing, and tape to the drawer. The red lines indicate the initial position variations of the objects in the collected 40 demonstrations.}
\label{fig:image4}
\vspace{-15pt}
\end{figure}

\paragraph{\textbf{System setup and task design}}
We conduct real-world experiments on 4 tasks with a Diana 7 robot arm and one RealSense D415 camera. As shown in figure \ref{fig:image4}, our tasks are (1) Pouring: grasp the bowl and pour the contents of the bowl into the pan; (2) Bottle\_upright: stand the horizontal bottle upright; (3) Tape\_to\_drawer: put the tape into the drawer and close the drawer; (4)Brushing: brush the pan with oil. We use GPT-4 \cite{GPT4} at the deployment stage to extract policy, target objects, and obstacles from user instruction and generate code to run the policy.


\vspace{3pt}

\paragraph{\textbf{Demonstrations collection}}
Demonstrations are collected via tele-operation with a space mouse and keyboard. We collect 40 demonstrations for each task with position variations shown by the red lines in figure \ref{fig:image4}. In the bottle upright task, the orientation of the bottle is not changed. Given the names of task-related objects, we invoke GroundingDINO \cite{GroundingDino} to predict the bounding boxes, Segment Anything \cite{SAM} to obtain the segmentation masks, and finally track the masks using video tracker Cutie \cite{Cutie}. We record demonstrations consisting of object point clouds and robot states, including the robot end-effector poses and gripper states, and the corresponding robot end-effector trajectories.

\subsubsection{Generalization evaluation}
\begin{figure*}[ht]
\centering
\vspace{6pt}
\includegraphics[width=0.99\textwidth]{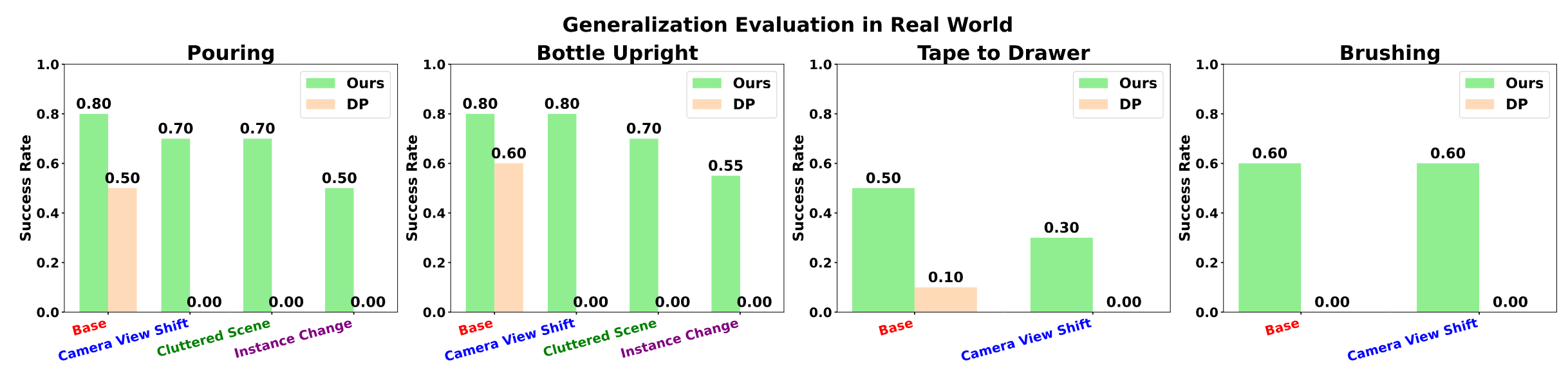}
\caption{Quantitative results of evaluating generalization. Our method has high success rates and strong generalization capability. Diffusion policy \cite{chi2023diffusionpolicy} achieves bad results because of limited demonstrations and poor generalization capabilities.}
\label{fig:image5}
\vspace{-15pt}
\end{figure*}

We evaluate the basic success rate of four tasks and generalization in challenging unseen environments compared with a baseline Diffusion Policy \cite{chi2023diffusionpolicy}. Figure \ref{fig:image5} shows that our approach demonstrates a higher success rate than the baseline and has significantly stronger generalization in unseen situations. Diffusion policy achieves bad results due to limited demonstrations and poor generalization capabilities with RGB inputs.

\begin{figure}[t]
\centering
\includegraphics[width=0.48\textwidth]{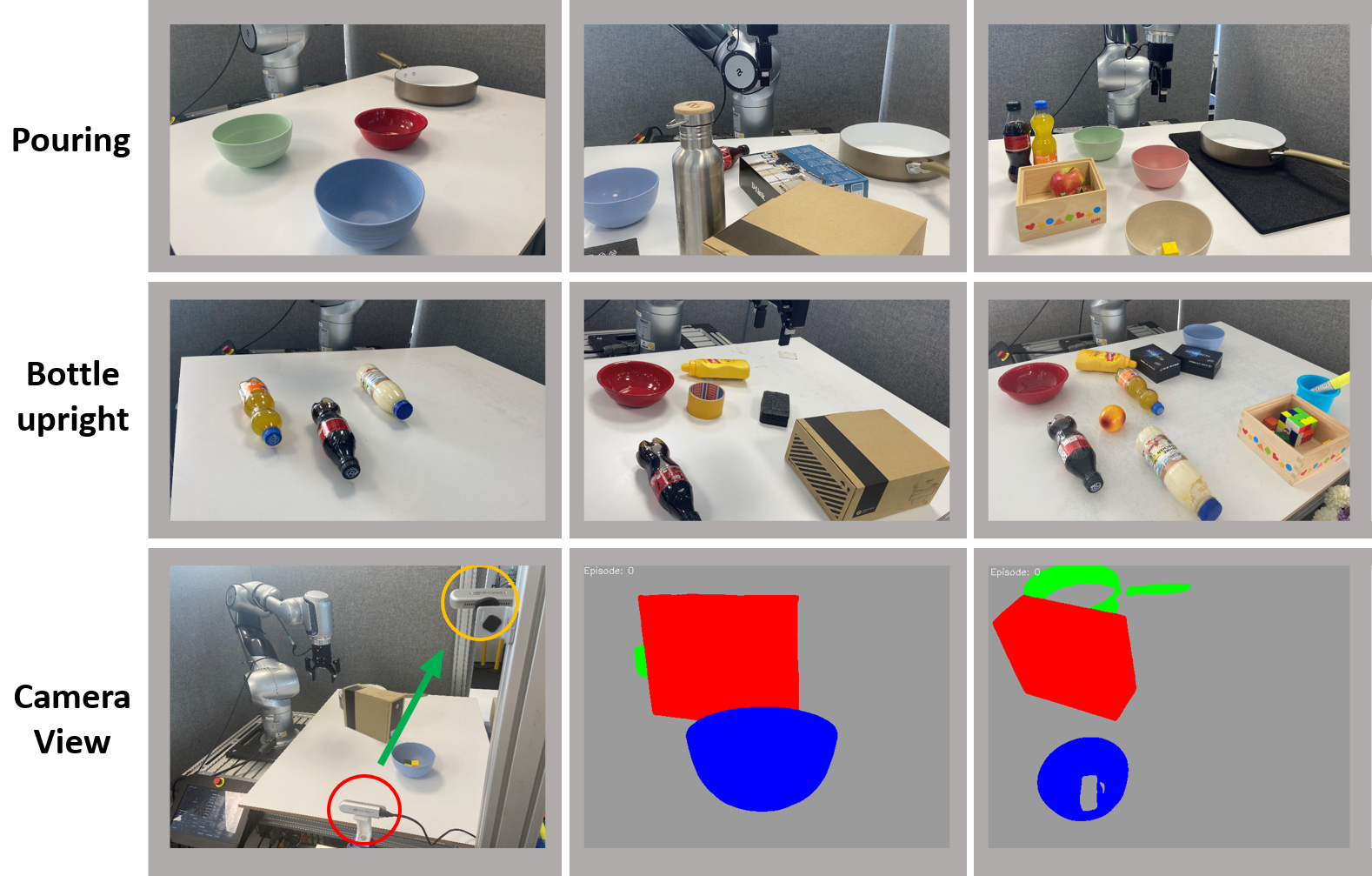}
\caption{Qualitative samples of testing scenes (instance changes, cluttered scene, multiple similar objects, camera view shift) for generalization evaluation.}
\label{fig:image6}
\vspace{-15pt}
\end{figure}

We evaluate the following aspects, and the figure \ref{fig:image6} shows some testing scenes.
(1) \textbf{Instance changes.}
We evaluate the generalization ability of objects with similar geometry through two tasks: bottle upright and pouring. We ask the robot to grasp the objects that are different from those in training. Our framework shows the ability to perform open-set manipulation with similar geometries. 
(2) \textbf{Multiple similar objects.}
In scenarios where multiple similar objects could be the target, we validate the importance of open-set segmentation and large language models. The policy can be effectively executed by specifying a particular object as the target using natural language. Our prompt for LLM is adapted from \cite{huang2023voxposer}.
(3) \textbf{Cluttered Scenes.}
In cluttered scenes, the challenge arises from the visual complexity and occlusion of the environment. Object-centric representation is only affected when the target is fully occluded, while the flexible camera view helps reveal the target.
(4) \textbf{Camera view shift.}
As shown in figure \ref{fig:image6}, in the third row, we only use the camera in the red circle for demonstration collection, and the camera in the orange circle is only for testing. In real-world experiments, our method has no performance drop when the camera changes from a red circle to an orange circle, while the RGB-based diffusion policy failed entirely. We found that a flexible camera view helps when the target objects are occluded in cluttered or obstacle-present scenes.

\vspace{3pt}

\subsubsection{Obstacle avoidance evaluation}
\begin{figure}[t]
\centering
\includegraphics[width=0.48\textwidth]{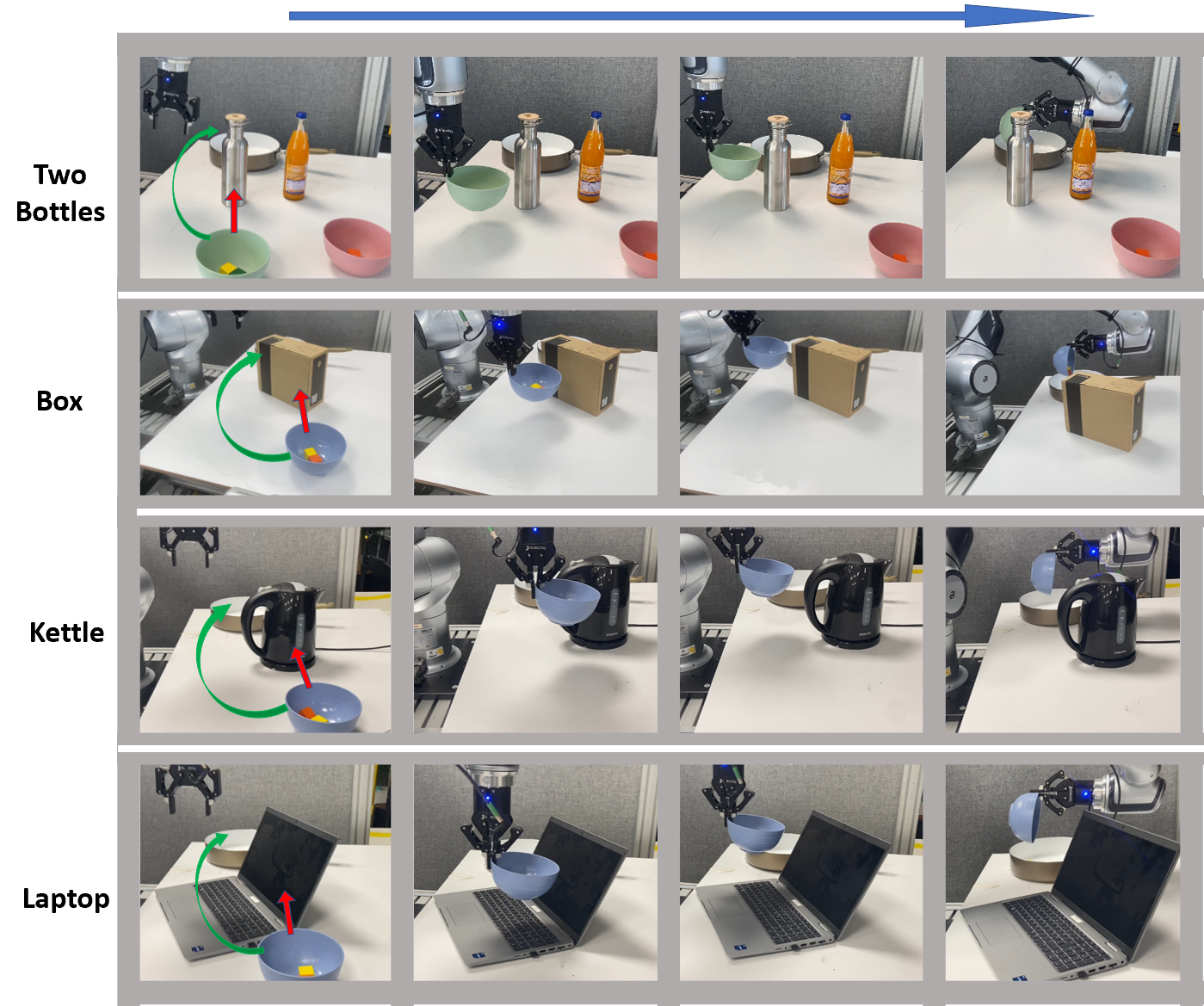}
\caption{Qualitative samples of open-set obstacle avoidance for objects of different shapes. Without guidance, the end-effector traces the path indicated by the red arrows and collides with the obstacles. With guidance, the path changes to the green arc, avoiding the obstacle.}
\label{fig:image7}
\vspace{-5pt}
\end{figure}

\begin{figure}[h]
\centering
\includegraphics[width=0.48\textwidth]{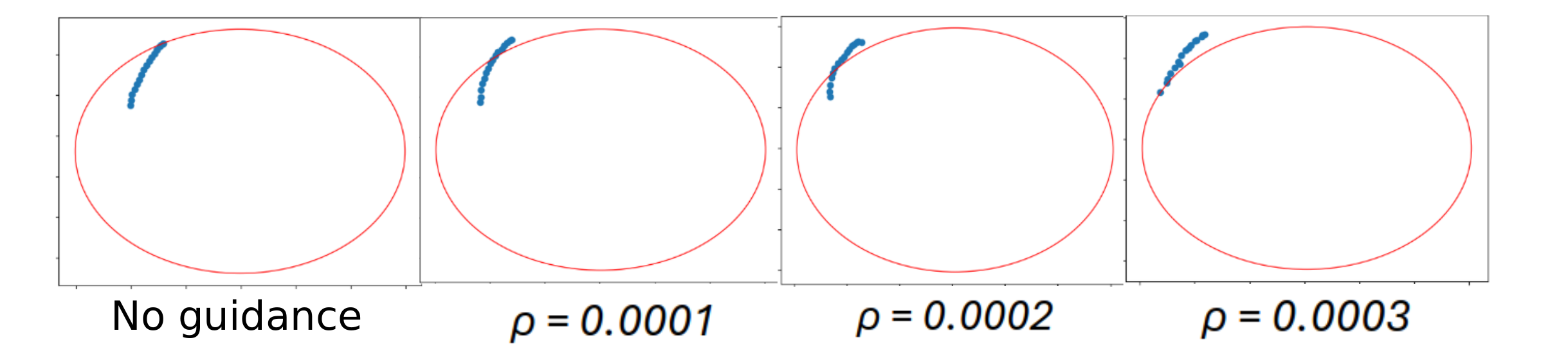}
\caption{Effect of gradient scale $\rho$ in DDPM 100 steps. Red circle indicates obstacles and blue points are generated waypoints.}
\label{fig:image8}
\vspace{-20pt}
\end{figure}

We test the open-set zero-shot obstacle avoidance in the pouring, brushing, and tape\_to\_drawer tasks with various obstacles such as bottles, a box, a laptop, a kettle, and a phone. The table \ref{obstacle} is the quantitative results with 5 trials conducted for each obstacle. We count failures when the robot or the grasped object hits the obstacles. Figure \ref{fig:image7} shows some qualitative rollouts of open-set obstacle avoidance. Our method is computationally efficient and can still maintain the original frequency, with 100 steps at approximately 1 Hz and 16 steps at around 5 Hz tested on RTX A6000 Ada. The gradient scale trades off the trajectory smoothness and the effectiveness of obstacle avoidance. Figure \ref{fig:image8} shows the effect of gradient scales.
\vspace{-10pt}
\begin{table}[htbp]
\centering
\caption{Quantitative results of obstacle avoidance}
\resizebox{\columnwidth}{!}{%
\begin{tabular}{@{}cccccccc@{}}
\toprule
 & \multicolumn{5}{c}{Pour} & Brush & Item in drawer \\
\cmidrule(lr){2-6} \cmidrule(lr){7-7} \cmidrule(lr){8-8}
 & Bottles & Box & Kettle & Phone & Laptop & Bottles & Bottles \\
\midrule
w/o guidance & 0\% & 0\% & 0\% & 0\% & 0\% & 0\% & 0\% \\
Ours & \textbf{100\%} & \textbf{60\%} & \textbf{80\%} & \textbf{100\%} & \textbf{60\%} & \textbf{40\%} & \textbf{20\%} \\
\bottomrule
\end{tabular}%
}
\label{obstacle}
\vspace{-10pt}
\end{table}

\textbf{Effectiveness of cost on $\mathbf{A_{0|k}}$.} Our proposed guidance achieves more efficient guidance and enables obstacle avoidance within 20 steps in DDIM by increasing the gradient scale. We found that the baseline method on $A_{k}$ only contributes to the final denoising step for the diffusion policy conditioned on observations, making it more akin to post-processing. Table \ref{Ak} shows the results of obstacle avoidance on the pouring task when removing the guidance on the final denoising step with 5 trials conducted for each obstacle.

\begin{table}[htbp]
\vspace{-10pt}
\centering
\caption{Quantitative results w/o the last step guidance}
\begin{tabular}{@{}cccc@{}}
\toprule
 & Bottles & Box & Kettle \\
\midrule
$A_{k}$   & 0\% & 0\% & 0\% \\
$A_{0|k}$ & \textbf{100\%} & \textbf{60\%} & \textbf{80\%} \\
\bottomrule
\end{tabular}
\label{Ak}
\vspace{-10pt}
\end{table}

\section{CONCLUSION}
\vspace{-5pt}
In this work, we propose a language-guided object-centric diffusion policy framework that is generalizable, collision-aware and data efficient. We use the point clouds of the target objects as the input for the diffusion policy to enhance the generalization and data efficiency. We introduced cost-guided trajectory generation for training-free obstacle avoidance. This work still has some limitations. We assume the target objects can be successfully detected and segmented, while the performance of the current VLM is limited. Additionally, our method can not be used for whole-body obstacle avoidance but only end-effector. Future work could involve using more advanced vision language models, incorporating LLM-based task planning, and designing cost functions on configuration space for whole-body collision avoidance. 

\balance
\bibliographystyle{IEEEtran}
\bibliography{chapter/reference}

\begin{thebibliography}{10}
\providecommand{\url}[1]{#1}
\csname url@samestyle\endcsname
\providecommand{\newblock}{\relax}
\providecommand{\bibinfo}[2]{#2}
\providecommand{\BIBentrySTDinterwordspacing}{\spaceskip=0pt\relax}
\providecommand{\BIBentryALTinterwordstretchfactor}{4}
\providecommand{\BIBentryALTinterwordspacing}{\spaceskip=\fontdimen2\font plus
\BIBentryALTinterwordstretchfactor\fontdimen3\font minus \fontdimen4\font\relax}
\providecommand{\BIBforeignlanguage}[2]{{%
\expandafter\ifx\csname l@#1\endcsname\relax
\typeout{** WARNING: IEEEtran.bst: No hyphenation pattern has been}%
\typeout{** loaded for the language `#1'. Using the pattern for}%
\typeout{** the default language instead.}%
\else
\language=\csname l@#1\endcsname
\fi
#2}}
\providecommand{\BIBdecl}{\relax}
\BIBdecl

\bibitem{brohan2023rt2}
A.~Brohan, N.~Brown, J.~Carbajal, Y.~Chebotar, X.~Chen, and et.al, ``Rt-2: Vision-language-action models transfer web knowledge to robotic control,'' in \emph{Proceedings of International Conference on Computer Vision}, 2023.

\bibitem{openvla}
M.~J. Kim, K.~Pertsch, S.~Karamcheti, T.~Xiao, A.~Balakrishna, S.~Nair, R.~Rafailov, E.~Foster, G.~L.~P. Sanketi, Q.~Vuong, T.~Kollar, B.~Burchfiel, R.~Tedrake, D.~Sadigh, S.~Levine, P.~Liang, and C.~Finn, ``Openvla: An open-source vision-language-action model,'' in \emph{Proceedings of Conference on Robot Learning(CoRL)}, 2024.

\bibitem{janner2022planning}
M.~Janner, Y.~Du, J.~B. Tenenbaum, and S.~Levine, ``Planning with diffusion for flexible behavior synthesis,'' in \emph{Proceedings of International Conference on Machine Learning (ICML)}, 2022.

\bibitem{ma2024hierarchical}
X.~Ma, S.~Patidar, I.~Haughton, and S.~James, ``Hierarchical diffusion policy for kinematics-aware multi-task robotic manipulation,'' in \emph{Proceedings of IEEE / CVF Computer Vision and Pattern Recognition Conference (CVPR)}, 2024.

\bibitem{gdp}
M.~Reuss, M.~X. Li, X.~Jia, and R.~Lioutikov, ``Goal conditioned imitation learning using score-based diffusion policies,'' in \emph{Proceedings of Robotics: Science and Systems (RSS)}, 2023.

\bibitem{IHB}
T.~Pearce, T.~Rashid, A.~Kanervisto, D.~Bignell, M.~Sun, R.~Georgescu, S.~V. Macua, S.~Z. Tan, I.~Momennejad, K.~Hofmann, and S.~Devlin, ``Imitating human behaviour with diffusion models,'' in \emph{Proceedings of ICLR}, 2023.

\bibitem{chi2023diffusionpolicy}
C.~Chi, S.~Feng, Y.~Du, Z.~Xu, E.~Cousineau, B.~Burchfiel, and S.~Song, ``Diffusion policy: Visuomotor policy learning via action diffusion,'' in \emph{Proceedings of Robotics: Science and Systems (RSS)}, 2023.

\bibitem{Ze2024DP3}
Y.~Ze, G.~Zhang, K.~Zhang, C.~Hu, M.~Wang, and H.~Xu, ``3d diffusion policy: Generalizable visuomotor policy learning via simple 3d representations,'' in \emph{Proceedings of Robotics: Science and Systems (RSS)}, 2024.

\bibitem{GroundingDino}
S.~Liu, Z.~Zeng, T.~Ren, F.~Li, J.~Y. Hao~Zhang, C.~Li, J.~Yang, H.~Su, J.~Zhu, and L.~Zhang, ``Grounding dino: Marrying dino with grounded pre-training for open-set object detection,'' 2023.

\bibitem{SAM}
A.~Kirillov, E.~Mintun, N.~Ravi, H.~Mao, L.~G. C.~Rolland, T.~Xiao, S.~Whitehead, A.~C. Berg, W.-Y. Lo, and et~al., ``Segment anything,'' in \emph{Proceedings of International Conference on Computer Vision (ICCV)}, 2023.

\bibitem{Cutie}
H.~K. Cheng, S.~W. Oh, B.~Price, J.-Y. Lee, and A.~Schwing, ``Putting the object back into video object segmentation,'' in \emph{Proceedings of Computer Vision and Pattern Recognition Conference (CVPR)}, 2024.

\bibitem{james2019rlbench}
S.~James, Z.~Ma, D.~R. Arrojo, and A.~J. Davison, ``Rlbench: The robot learning benchmark \& learning environment,'' 2019.

\bibitem{Liang_2019}
\BIBentryALTinterwordspacing
H.~Liang, X.~Ma, S.~Li, M.~Gorner, S.~Tang, B.~Fang, F.~Sun, and J.~Zhang, ``Pointnetgpd: Detecting grasp configurations from point sets,'' in \emph{2019 International Conference on Robotics and Automation (ICRA)}.\hskip 1em plus 0.5em minus 0.4em\relax IEEE, May 2019. [Online]. Available: \url{http://dx.doi.org/10.1109/ICRA.2019.8794435}
\BIBentrySTDinterwordspacing

\bibitem{Qian2022FFHNet}
\BIBentryALTinterwordspacing
V.~Mayer*, Q.~Feng*, J.~Deng, Y.~Shi, Z.~Chen, and A.~Knoll, ``Ffhnet: Generating multi-fingered robotic grasps for unknown objects in real-time,'' \emph{2022 International Conference on Robotics and Automation (ICRA)}, pp. 762--769, 2022. [Online]. Available: \url{https://api.semanticscholar.org/CorpusID:250508500}
\BIBentrySTDinterwordspacing

\bibitem{burkhardt2024multifingered}
Y.~Burkhardt*, Q.~Feng*, J.~Feng, K.~Sharma, Z.~Chen, and A.~Knoll, ``Multi-fingered dynamic grasping for unknown objects,'' 2024.

\bibitem{feng2024dexgangrasp}
Q.~Feng, D.~S.~M. Lema, M.~Malmir, H.~Li, J.~Feng, Z.~Chen, and A.~Knoll, ``Dexgangrasp: Dexterous generative adversarial grasping synthesis for task-oriented manipulation,'' 2024.

\bibitem{zhu2023groot}
Y.~Zhu, Z.~Jiang, P.~Stone, and Y.~Zhu, ``Learning generalizable manipulation policies with object-centric 3d representations,'' in \emph{Proceedings of Conference on Robot Learning (CoRL)}, 2023.

\bibitem{liang2024skilldiffuser}
Z.~Liang, Y.~Mu, H.~Ma, M.~Tomizuka, M.~Ding, and P.~Luo, ``Skilldiffuser: Interpretable hierarchical planning via skill abstractions in diffusion-based task execution,'' in \emph{Proceedings of IEEE / CVF Computer Vision and Pattern Recognition Conference (CVPR)}, 2024.

\bibitem{Bimanual}
T.~B. Akbulut, A.~M. G.~Tuba C.~Girgin, M.~Asada, E.~Ugur, and E.~Oztop, ``Bimanual rope manipulation skill synthesis through context dependent correction policy learning from human demonstration,'' in \emph{Proceedings of International Conference on Robotics and Automation(ICRA)}, 2023.

\bibitem{Surgical}
D.~Zhang, Z.~Wu, J.~Chen, R.~Zhu, A.~Munawar, B.~Xiao, Y.~Guan, H.~Su, W.~Hong, Y.~Guo, G.~S. Fischer, B.~Lo, and G.-Z. Yang, ``Human-robot shared control for surgical robot based on context-aware sim-to-real adaptation,'' in \emph{Proceedings of International Conference on Robotics and Automation(ICRA)}, 2022.

\bibitem{ajay2023conditional}
A.~Ajay, Y.~Du, A.~Gupta, J.~Tenenbaum, T.~Jaakkola, and P.~Agrawal, ``Is conditional generative modeling all you need for decision-making?'' in \emph{Proceedings of International Conference on Learning Representations (ICLR)}, 2023.

\bibitem{adaptdiffuser}
Z.~Liang, Y.~Mu, M.~Ding, F.~Ni, M.~Tomizuka, and P.~Luo, ``Adaptdiffuser: Diffusion models as adaptive self-evolving planners,'' in \emph{Proceedings of International Conference on Machine Learning(ICML)}, 2023.

\bibitem{barad2023graspldm}
K.~R. Barad, A.~Orsula, A.~Richard, J.~Dentler, M.~Olivares-Mendez, and C.~Martinez, ``Graspldm: Generative 6-dof grasp synthesis using latent diffusion models,'' 2023.

\bibitem{weng2024dexdiffuser}
Z.~Weng, H.~Lu, D.~Kragic, and J.~Lundell, ``Dexdiffuser: Generating dexterous grasps with diffusion models,'' 2024.

\bibitem{urain2023se3diffusionfields}
J.~Urain, N.~Funk, J.~Peters, and G.~Chalvatzaki, ``Se(3)-diffusionfields: Learning smooth cost functions for joint grasp and motion optimization through diffusion,'' in \emph{2023 International Conference on Robotics and Automation (ICRA)}.\hskip 1em plus 0.5em minus 0.4em\relax IEEE, 2023.

\bibitem{MPD}
J.~Carvalho, A.~T. Le, M.~Baierl, D.~Koert, and J.~Peters, ``Motion planning diffusion: Learning and planning of robot motions with diffusion models,'' in \emph{Proceedings of International Conference on Intelligent Robots and Systems (IROS)}, 2023.

\bibitem{EDMP}
K.~Saha, V.~Mandadi, J.~Reddy, A.~Srikanth1, A.~Agarwal, B.~Sen, A.~Singh, and M.~Krishna1, ``Ensemble-of-costs-guided diffusion for motion planning,'' in \emph{Proceedings of International Conference on Robotics and Automation(ICRA)}, 2024.

\bibitem{pose1}
\BIBentryALTinterwordspacing
J.~Tremblay, T.~To, B.~Sundaralingam, Y.~Xiang, D.~Fox, and S.~Birchfield, ``Deep object pose estimation for semantic robotic grasping of household objects,'' \emph{CoRR}, vol. abs/1809.10790, 2018. [Online]. Available: \url{http://arxiv.org/abs/1809.10790}
\BIBentrySTDinterwordspacing

\bibitem{pose2}
S.~Tyree, J.~Tremblay, T.~To, J.~Cheng, T.~Mosier, J.~Smith, and S.~Birchfield, ``6-dof pose estimation of household objects for robotic manipulation: An accessible dataset and benchmark,'' 2022.

\bibitem{pose3}
\BIBentryALTinterwordspacing
T.~Migimatsu and J.~Bohg, ``Object-centric task and motion planning in dynamic environments,'' \emph{CoRR}, vol. abs/1911.04679, 2019. [Online]. Available: \url{http://arxiv.org/abs/1911.04679}
\BIBentrySTDinterwordspacing

\bibitem{wang2019deep}
D.~Wang, C.~Devin, Q.-Z. Cai, F.~Yu, and T.~Darrell, ``Deep object-centric policies for autonomous driving,'' 2019.

\bibitem{devin2017deep}
C.~Devin, P.~Abbeel, T.~Darrell, and S.~Levine, ``Deep object-centric representations for generalizable robot learning,'' 2017.

\bibitem{sdmaskrcnn}
\BIBentryALTinterwordspacing
M.~Danielczuk, M.~Matl, S.~Gupta, A.~Li, A.~Lee, J.~Mahler, and K.~Goldberg, ``Segmenting unknown 3d objects from real depth images using mask {R-CNN} trained on synthetic point clouds,'' \emph{CoRR}, vol. abs/1809.05825, 2018. [Online]. Available: \url{http://arxiv.org/abs/1809.05825}
\BIBentrySTDinterwordspacing

\bibitem{cheng2022xmem}
H.~K. Cheng and A.~G. Schwing, ``Xmem: Long-term video object segmentation with an atkinson-shiffrin memory model,'' in \emph{Proceedings of European Conference on Computer Vision (ECCV)}, 2022.

\bibitem{CodeasPolicies}
J.~Liang, W.~Huang, F.~Xia, P.~Xu, K.~Hausman, B.~Ichter, P.~Florence, A.~Zeng, and et.al, ``Code as policies: Language model programs for embodied control,'' in \emph{IEEE International Conference on Robotics and Automation (ICRA)}, 2023.

\bibitem{huang2023voxposer}
W.~Huang, C.~Wang, R.~Zhang, Y.~Li, J.~Wu, and L.~Fei-Fei, ``Voxposer: Composable 3d value maps for robotic manipulation with language models,'' in \emph{Proceedings of Conference on Robot Learning(CoRL)}, 2023.

\bibitem{saycan}
M.~Ahn, A.~Brohan, N.~Brown, Y.~Chebotar, O.~Cortes, and et.al, ``Do as i can, not as i say: Grounding language in robotic affordances,'' in \emph{Proceedings of Conference on Robot Learning(CoRL)}, 2022.

\bibitem{reward1}
W.~Yu, N.~Gileadi, C.~Fu, S.~Kirmani, K.-H. Lee, and et.al, ``Language to rewards for robotic skill synthesis,'' in \emph{Proceedings of Conference on Robot Learning(CoRL)}, 2023.

\bibitem{reward2}
T.~Xie, S.~Zhao, C.~H. Wu, Y.~Liu, Q.~Luo, V.~Zhong, Y.~Yang, and T.~Yu, ``Text2reward: Reward shaping with language models for reinforcement learning,'' in \emph{Proceedings of International Conference on Learning Representations(ICLR)}, 2024.

\bibitem{Eureka}
J.~Ma, W.~Liang, G.~Wang, D.-A. Huang, O.~Bastani, D.~Jayaraman, Y.~Zhu, L.~J. Fan, and A.~Anandkumar1, ``Eureka: Human-level reward design via coding large language models,'' in \emph{Proceedings of International Conference on Learning Representations(ICLR)}, 2024.

\bibitem{OWL-ViT}
M.~Minderer, A.~Gritsenko, A.~Stone, M.~Neumann, and et.al, ``Simple open-vocabulary object detection with vision transformers,'' in \emph{Proceedings of European Conference on Computer Vision (ECCV)}, 2022.

\bibitem{yolo-world}
T.~Cheng, L.~Song, Y.~Ge, W.~Liu, X.~Wang, and Y.~Shan, ``Yolo-world: Real-time open-vocabulary object detection,'' in \emph{Proceedings of Computer Vision and Pattern Recognition Conference (CVPR)}, 2024.

\bibitem{stone2023openworld}
A.~Stone, T.~Xiao, Y.~Lu, K.~Gopalakrishnan, K.-H. Lee, Q.~Vuong, P.~Wohlhart, S.~Kirmani, B.~Zitkovich, F.~Xia, C.~Finn, and K.~Hausman, ``Open-world object manipulation using pre-trained vision-language models,'' in \emph{Proceedings of Conference on Robot Learning(CoRL)}, 2023.

\bibitem{FiLM}
E.~Perez, F.~Strub, H.~de~Vries, V.~Dumoulin, and A.~Courville, ``Film: Visual reasoning with a general conditioning layer,'' in \emph{Proceedings of Conference on Artificial Intelligence(AAAI)}, 2018.

\bibitem{FreeDoM}
J.~Yu, Y.~Wang, C.~Zhao, B.~Ghanem, and J.~Zhang, ``Freedom: Training-free energy-guided conditional diffusion model,'' in \emph{Proceedings of International Conference on Computer Vision}, 2023.

\bibitem{DDPM}
J.~Ho, A.~Jain, and P.~Abbeel, ``Denoising diffusion probabilistic models,'' in \emph{Proceedings of NeurIPS}, 2020.

\bibitem{DDIM}
J.~Song, C.~Meng, and S.~Ermon, ``Denoising diffusion implicit models,'' in \emph{Proceedings of ICLR}, 2021.

\bibitem{pointnet}
C.~R. Qi, H.~Su, K.~Mo, and L.~J. Guibas, ``Pointnet: Deep learning on point sets for 3d classification and segmentation,'' in \emph{Proceedings of Computer Vision and Pattern Recognition Conference (CVPR)}, 2017.

\bibitem{GPT4}
OpenAI, J.~Achiam, S.~Adler, S.~Agarwal, L.~Ahmad, and et.al, ``Gpt-4 technical report,'' 2023.

\end{thebibliography}

\end{document}